\documentclass{article}
\usepackage[final]{neurips_2025}
\usepackage[utf8]{inputenc} % allow utf-8 input
\usepackage[T1]{fontenc}    % use 8-bit T1 fonts
\usepackage{hyperref}       % hyperlinks
\usepackage{url}            % simple URL typesetting
\usepackage{booktabs}       % professional-quality tables
\usepackage{amsfonts}       % blackboard math symbols
\usepackage{nicefrac}       % compact symbols for 1/2, etc.
\usepackage{microtype}      % microtypography
\usepackage{graphicx}       % for including images
\usepackage{xcolor}         % for text color
\usepackage[utf8]{inputenc} % allow utf-8 input
\usepackage[T1]{fontenc}    % use 8-bit T1 fonts
\usepackage{hyperref}       % hyperlinks
\usepackage{url}            % simple URL typesetting
\usepackage{booktabs}       % professional-quality tables
\usepackage{amsfonts}       % blackboard math symbols
\usepackage{nicefrac}       % compact symbols for 1/2, etc.
\usepackage{microtype}      % microtypography
\usepackage{graphicx}       % for including images
\usepackage{xcolor}         % for text color
\usepackage{filecontents}   % To embed the .bib file
\usepackage{listings}
\title{Operation Veja: Fixing Fundamental Concepts Missing from Modern Roleplaying Training Paradigms}

\author{%
  Yueze Liu \\
  Divergence 2\% LLC \\
  \texttt{lolingandtroll@gmail.com} \\
  \And
  Ajay Nagi Reddy Kumdam \\
  Department of Electrical and Computer Engineering \\
  University of Illinois Urbana-Champaign \\
  \And
  Ronit Kanjilal \quad Hao Yang \quad Yichi Zhang \\
  Department of Computer Science \\
  University of Illinois Urbana-Champaign \\
}

\begin{document}

\maketitle

% --- Abstract ---
\begin{abstract}
Modern roleplaying models are increasingly sophisticated, yet they consistently fail to capture the essence of believable, engaging characters. We argue this failure stems from training paradigms that overlook the dynamic interplay of a character's internal world. Current approaches, including Retrieval-Augmented Generation (RAG), fact-based priming, literature-based learning, and synthetic data generation, are fundamentally limited in their ability to model the deliberative, value-conflicted reasoning that defines human interaction. In this paper, we identify four core classic concepts essential for character authenticity: Values, Experiences, Judgments, and Abilities (VEJA). We propose the VEJA framework as a new paradigm for data curation that addresses these systemic failures. To illustrate the qualitative ceiling that our framework enables, we present a pilot study comparing a manually-curated, VEJA-grounded dataset against a state-of-the-art synthetic baseline. Using an LLM-as-judge, our findings demonstrate a significant quality gap, suggesting that a fundamental shift towards conceptually-grounded data curation, as embodied by VEJA, is necessary for creating roleplaying agents with genuine depth and narrative continuity. The full dataset is available at \url{https://github.com/HyouinKyoumaIRL/Operation-Veja}.

\end{abstract}

% --- Introduction ---
\section{Introduction}

Modern roleplaying models are highly sought after, whether for creating immersive gaming experiences or for developing persistent, captivating AI companions. Pioneering work in creating interactive simulacra of human behavior has demonstrated complex agentic systems \citep{park2023generative}, yet the underlying characters often lack the deliberative internal reasoning we associate with genuine personality. The primary function of a roleplaying agent is to assume a personality, acting with a predefined set of values, capabilities, and experiences, and to infer these traits when not explicitly provided.

However, we argue that current models fail to capture the most critical element of a believable character: the dynamic interplay of a complex internal value system. Human interaction is not a process of retrieving the single "correct" response. It is a constant negotiation of conflicting values—the balance between being polite versus efficient, expressing curiosity versus maintaining practicality, or being vulnerable versus appearing professional. Today's models lack this internal, deliberative process, resulting in characters that feel shallow, predictable, and ultimately, unconvincing. When we first attempted to construct a high-fidelity model of the character Makise Kurisu from \textit{Steins;Gate}, we found that no existing technique could replicate her core trait of intellectual curiosity clashing with social guardedness; responses were merely reactionary, never truly inquisitive or driven by an internal conflict.

This paper aims to address this fundamental gap. We will:
\begin{enumerate}
    \item Critically analyze the systemic failures of four dominant approaches to character modeling.
    \item Propose the VEJA framework (Values, Experiences, Judgments, Abilities) as the needed baseline for simulating character depth and ensuring narrative continuity.
    \item Present preliminary results from a small-scale, manually-curated dataset that demonstrates the promise of our approach, highlighting the need for a community-wide shift in how we create and evaluate roleplaying agents.
\end{enumerate}

% --- Critical Analysis ---
\section{A Critical Analysis of Current Roleplaying Paradigms}
\label{sec:analysis}

We identify four primary methodologies used in training roleplaying models, each with inherent limitations that prevent the emergence of authentic character depth.

\subsection{Retrieval-Augmented Generation (RAG): The Scaling Problem}
RAG attempts to ground a model in a character's "facts" by retrieving from a knowledge base \citep{lewis2020retrieval}. While useful for factual recall, this approach breaks down when applied to a character's internal state. The human value system is combinatorially explosive and too vast to enumerate. A character's views on digital media, anxiety marketing, monetary policy, and marriage are not discrete facts. They are interconnected and context-dependent. When a situational modifier is introduced—for instance, "this person is attractive, so I will temporarily loosen my professional judgments"—the complexity grows beyond simply choosing the "correct" value. This makes the RAG approach intractable for modeling the nuanced, higher-plexity reasoning that defines personality.

\subsection{Fact-Based Value Priming: The Decontextualization Problem}
This method is a direct consequence of an evaluation landscape focused on measurable "personality fidelity" \citep{wang2024incharacter} through benchmarks like RoleLLM \citep{wang2024rolellm}. It involves generating lists of values and interviewing the model with questions like, "How would you act? What would you think?" The unfortunate result is that the model learns to represent values as simple, isolated formulas, stripped of context. For example, a character whose profile states they are "open with strangers" will often over-index on this single trait, bluntly stating their thoughts despite social cues that would dictate subtlety. This is a direct artifact of a data construction pipeline that exhibits values directly rather than embedding them within realistic, multi-faceted scenarios where they must be balanced against other competing values.

\subsection{Literature-Based Generation: The Curse of Implicit Context and Complexity}
Approaches like COSER \citep{wang2025coser} and ChatHaruhi \citep{li2023chatharuhi}, which use dialogue from literary masterpieces, seem promising. Characters in great novels possess the very depth and subtext we seek to emulate. However, this blessing is also a curse for two reasons.

First, the internal reasoning behind a character's dialogue is often more rigorous than what is explicitly stated. As even the authors of COSER note, their evaluation framework is designed to capture complex behaviors like inner thoughts and motivations \citep{wang2025coser}, yet the training data itself—the dialogue—can only ever *imply* the deliberative process: "What do I know? What do I want? Am I comfortable joking? Is this topic boring? How can I explain what I'm worried about?" While most are subconscious, missing even a few of these deliberative steps shatters believability. Literary text implies this process but does not provide it as explicit training data.

Second, literary works handle experiential context through narration, not dialogue. A character's foundational experiences, such as their childhood, are rarely conveyed through direct conversational recall in the training snippets. Consequently, the concept of experience as a precedent for current action is not learned. When a model is given a backstory, it often ignores it because its training data lacks examples of characters grounding their present-day dialogue in past events.

\subsection{Synthetic Data Generation and Distillation: The Chicken-and-Egg Problem}
The most alluring trap is the idea of using a powerful base model to generate value-grounded dialogue, an approach now being attempted at a massive scale \citep{chan2024scaling}. If a model understands all the nuances of conversation, it should be able to generate a perfect dataset. The fatal flaw is that no current model possesses this capability.

This creates a recursive quality ceiling. When a model like GPT-4 is asked to generate a "realistic first encounter" for a character with conflicting values—e.g., "I want to share my knowledge, but I also want to come across as professional and guarded"—it cannot effectively balance the conflict. The model will inevitably over-align with one value, leading to the brittle "overly mean" or "over-sharing" personas common in many state-of-the-art roleplaying bots. The very act of breaking character to resolve an internal conflict is, at present, an impossible task for a model to generate synthetically, because the generator itself lacks that capacity.

% --- VEJA Framework ---
\section{The VEJA Framework}
\label{sec:veja}

Our frustrations with these existing paradigms led us to develop a new framework for character definition, born from our data curation process. We found that the most jarring inconsistencies in writer-contributed data arose from a lack of grounding in four interconnected concepts. The VEJA framework provides this grounding. In doing so, it draws inspiration from long-established principles in dramatic arts, which have long understood that character is revealed through the interplay of objective, motivation, and past circumstance \citep{stanislavski2008actors}.

\begin{figure}[h]
  \centering
  \includegraphics[width=0.8\linewidth]{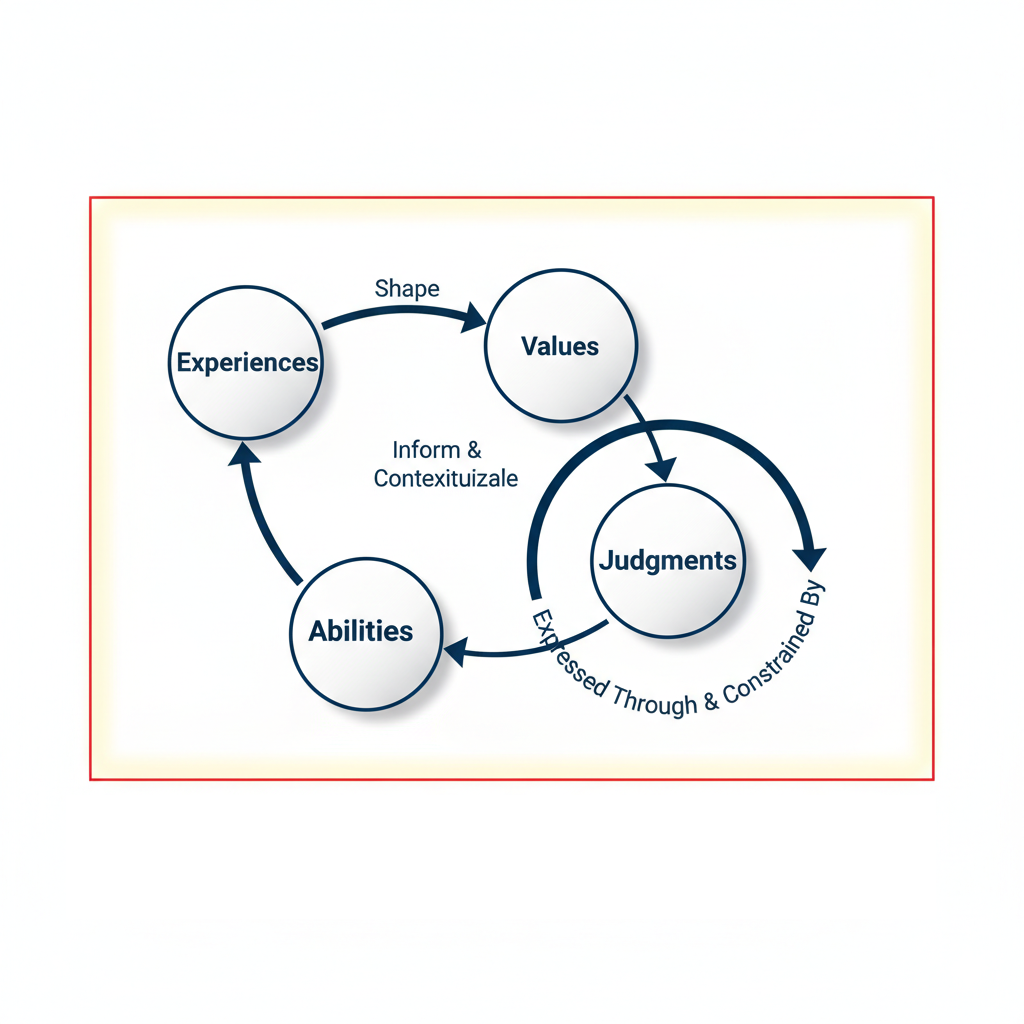}
  \caption{Conceptual cycle illustrating how Values, Experiences, Judgments, and Abilities interact. Experiences shape Values, which together inform Judgments, expressed through Abilities.}
  \label{fig:inner_cycle}
\end{figure}

\subsection{Values (V)}
Values are the fundamental, high-level motivators that drive a character's goals and desires. They are the "why" behind their actions.
\begin{itemize}
    \item \textbf{Example:} "I want more time because I have a lot I want to do" (Value: Ambition/Productivity).
    \item \textbf{Example:} "I want to be a minimalist because few things are truly important" (Value: Simplicity/Focus).
    \item \textbf{Example:} "I want to enjoy life to the fullest in the time I have" (Value: Hedonism/Experience-seeking).
\end{itemize}

\subsection{Experiences (E)}
Experiences are specific past events that shape and give texture to a character's Values and Judgments. They are the evidentiary basis for their worldview and provide the context for their emotional responses.
\begin{itemize}
    \item \textbf{Example:} "I am afraid of snakes because I was bitten as a child" (Experience grounds an emotional response).
    \item \textbf{Example:} "I went to court over a contract dispute, so now I am deeply skeptical of handshake deals" (Experience informs a Judgment).
    \item \textbf{Example:} "Reading subpar academic papers has made me critical of modern research trends" (Experience shapes a professional opinion).
\end{itemize}

\subsection{Judgments (J)}
Judgments are the concrete opinions, beliefs, and heuristics a character applies to the world. They are the outputs of Values being filtered through Experiences.
\begin{itemize}
    \item \textbf{Example:} "I think marriage is an outdated institution" (A specific belief).
    \item \textbf{Example:} "I think people take themselves too seriously" (A social heuristic).
    \item \textbf{Example:} "I find talking like a hipster to be fun and ironic" (A stylistic preference).
\end{itemize}

\subsection{Abilities (A)}
Abilities are the character's skills, knowledge, and competencies. They define the domains in which the character can act with authority and influence their interests, vocabulary, and the topics they gravitate towards.
\begin{itemize}
    \item \textbf{Example:} "I can design PCBs, so I am interested when others talk about hardware engineering."
    \item \textbf{Example:} "I have deep experience with machine learning, which informs my technical vocabulary and problem-solving approach."
\end{itemize}

These four pillars are not isolated; they form a causal chain. Experiences forge Values; Values and Experiences together produce Judgments; and Abilities provide the toolkit through which the character expresses these traits. By defining a character along these four axes, we create a coherent internal logic that can be used to generate consistent, deeply-grounded dialogue.

% --- Experiment Section ---
\section{Case Study: Curating a VEJA-Grounded Dataset}
\label{sec:experiment}

To demonstrate the feasibility of the VEJA framework and illustrate the qualitative differences in data produced when human authors are guided by its principles. This pilot study serves as a proof-of-concept for the type of nuanced data we argue is missing, rather than a definitive comparative analysis.

\begin{figure}[h]
  \centering
  \includegraphics[width=0.9\linewidth]{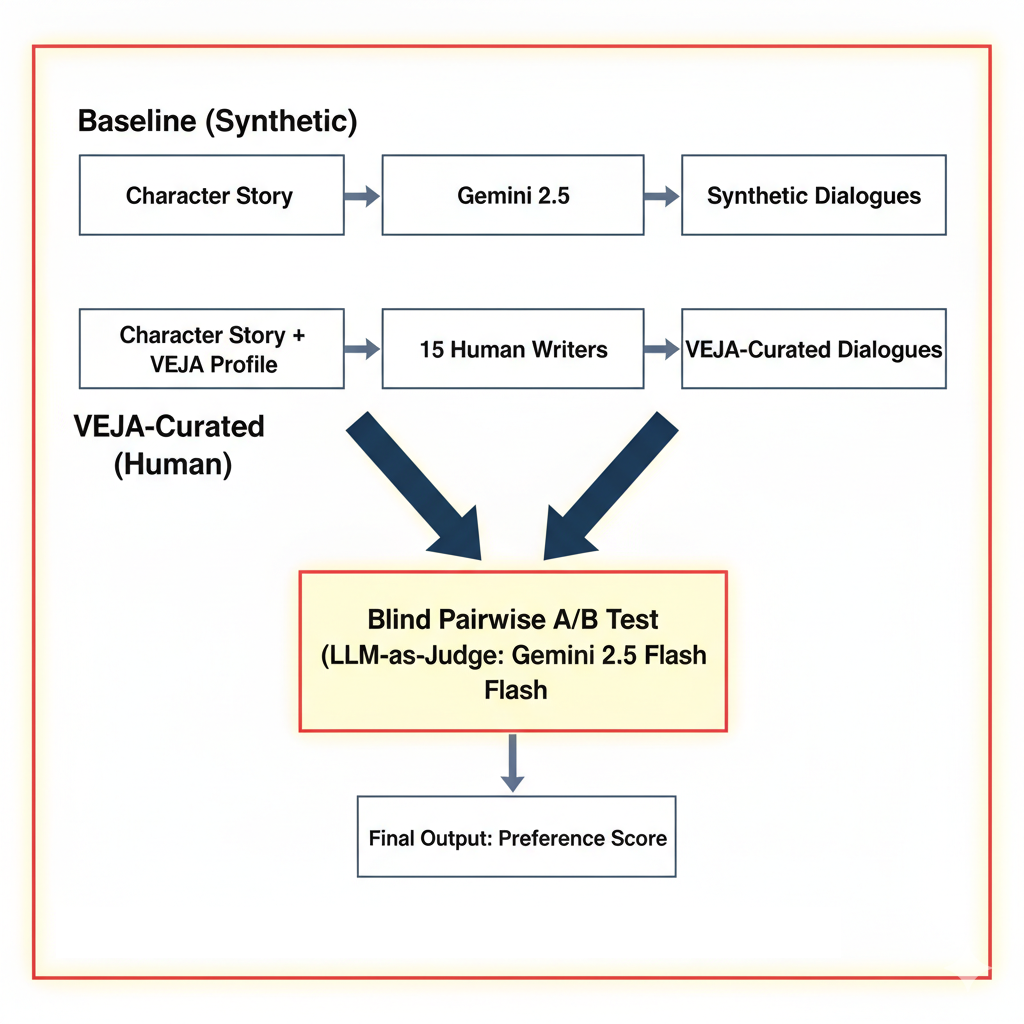}
  \caption{Experimental pipeline showing the baseline (top) and VEJA (bottom) branches, both feeding into the LLM-as-judge evaluation.}
  \label{fig:pipeline_diagram}
\end{figure}

\subsection{Dataset Curation}
We created two small, targeted datasets centered around the character Makise Kurisu.

\textbf{Baseline (Synthetic) Dataset:} We created a dictionary defining a 10-day story timeline for Makise Kurisu and 20 different users. We then prompted a large language model (\textit{Gemini Pro 2.5} \citep{comanici2025gemini}) to generate conversations between Kurisu and each user that were appropriate for the specified day and context.

\textbf{VEJA-Curated (Human) Dataset:} We provided 15 different human writers with the same 10-day story timeline for Kurisu. Crucially, they were also given a detailed character profile structured according to the VEJA framework. Each writer was tasked with imagining a conversation they would have with Kurisu, ensuring her dialogue and internal reasoning were grounded in the provided VEJA profile.

We explicitly chose to compare VEJA-guided human writing against a state-of-the-art synthetic baseline. We acknowledge that this introduces a human-vs-machine confound. However, our central argument is that current models cannot synthetically generate this level of quality, making skilled human curation—guided by a robust framework—an essential component for creating next-generation roleplaying data. This study therefore aims to establish a 'gold standard' quality ceiling that synthetic methods currently fail to reach.

\subsection{Evaluation Methodology}
We conducted a blind A/B test using an LLM-as-judge (Gemini 2.5 Flash), a common methodology with well-documented characteristics and potential biases \citep{zheng2023judging}. For each evaluation, we randomly sampled one conversation from the baseline dataset and one from the VEJA-curated dataset. The judge was presented with the two dialogues in a shuffled order (Dialogue A and Dialogue B) and was not informed of their origin. We performed 100 such pairwise comparisons. The judge's task was to determine which dialogue felt more like the character in generation, a proxy for naturalness and character consistency in this context. The full prompt is available in Appendix \ref{appendix:prompt}.

\subsection{Results}
The results demonstrated a strong preference for the data generated using the VEJA framework. Out of 100 pairwise comparisons, the VEJA-curated (human-written) dialogues were preferred 43 times, while the baseline (synthetic) dialogues were preferred 28 times, with 29 ties.

This corresponds to a significant win rate for the VEJA-curated dataset when ties are excluded. The LLM judge frequently noted that the preferred responses demonstrated superior narrative continuity, more nuanced reactions, and a clearer connection between the character's known personality and their dialogue. In contrast, the baseline synthetic data was often flagged as generic, out-of-character, or overly reliant on a single character trait. Example judgments are provided in Appendix \ref{appendix:examples}.

\begin{figure}[h]
  \centering
  \includegraphics[width=0.9\linewidth]{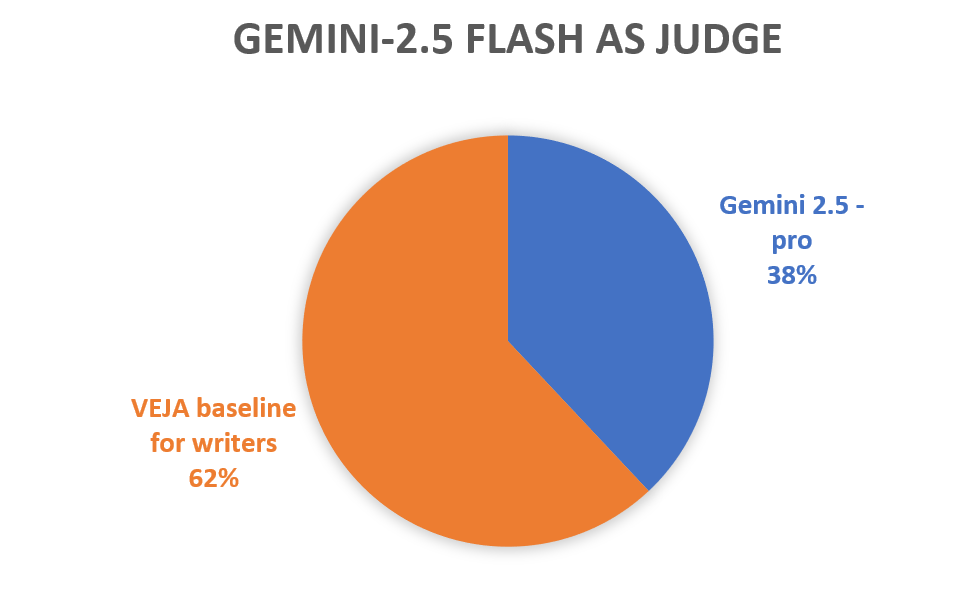}
  \caption{LLM-as-Judge Preference in Pairwise A/B Testing (N=100). The VEJA-curated dialogues were significantly preferred over the synthetically generated baseline.}
  \label{fig:pipeline_diagram}
\end{figure}

% --- Limitations ---
\section{Limitations}

Our study is preliminary and has several limitations. The dataset is small-scale and focuses on a single character archetype. The curation process was manual and labor-intensive, raising questions about scalability. Furthermore, our evaluation relies on an LLM-as-judge, which carries its own inherent biases \citep{zheng2023judging}. It is somewhat ironic that we critique synthetic data generation while using an LLM for evaluation; however, we believe its use in a comparative preference task is a valid, if imperfect, proxy for human judgment.

A significant confounder is that the preferred responses were human-written, while the baseline was machine-generated. We completely agree with this possibility. However, we contend that the VEJA framework was the essential tool that enabled the human writers to achieve such high consistency and quality. A core takeaway of this paper is that skilled human writers, equipped with the right conceptual tools, may be necessary for the next leap in LLM roleplaying capabilities. Due to time constraints, ablation studies analyzing the impact of individual VEJA components were not conducted.

% --- Conclusion and Future Work ---
\section{Conclusion and Future Work}

Current paradigms for training roleplaying models are hitting a ceiling. They produce characters that can mimic personality but cannot simulate it. By focusing on surface-level traits and isolated facts, they miss the interconnected web of values, experiences, judgments, and abilities that form the bedrock of a believable identity.

The VEJA framework is a first step toward a new approach—one centered on modeling a character's internal world. Our preliminary results are highly encouraging, suggesting that data curated with this deep, causal structure is significantly higher in quality.

Future work will focus on scaling the VEJA data creation process, potentially through semi-automated methods, and developing more robust evaluation metrics that go beyond preference scores to measure specific qualities like value-conflict resolution and experiential recall. Ultimately, the goal of Operation Veja is to shift the community's focus from building better parrots to creating truly authentic, deliberative, and unforgettable digital minds.

% --- References Section ---
\section*{References}
\bibliographystyle{plainnat}
\bibliography{references}

\begin{thebibliography}{10}
\providecommand{\natexlab}[1]{#1}
\providecommand{\url}[1]{\texttt{#1}}
\expandafter\ifx\csname urlstyle\endcsname\relax
  \providecommand{\doi}[1]{doi: #1}\else
  \providecommand{\doi}{doi: \begingroup \urlstyle{rm}\Url}\fi

\bibitem[Chan et~al.(2024)Chan, Wang, Yu, Mi, and Yu]{chan2024scaling}
Xin Chan, Xiaoyang Wang, Dian Yu, Haitao Mi, and Dong Yu.
\newblock Scaling synthetic data creation with 1,000,000,000 personas, 2024.

\bibitem[Comanici et~al.(2025)]{comanici2025gemini}
Gheorghe Comanici et~al.
\newblock Gemini 2.5: Pushing the frontier with advanced reasoning, multimodality, long context, and next generation agentic capabilities, 2025.

\bibitem[Lewis et~al.(2020)Lewis, Perez, Piktus, Petroni, Karpukhin, Goyal, K{\"u}ttler, Lewis, Yih, Kiela, et~al.]{lewis2020retrieval}
Patrick Lewis, Ethan Perez, Aleksandra Piktus, Fabio Petroni, Vladimir Karpukhin, Naman Goyal, Heinrich K{\"u}ttler, Mike Lewis, Wen-tau Yih, Douwe Kiela, et~al.
\newblock Retrieval-augmented generation for knowledge-intensive {NLP} tasks.
\newblock In \emph{Advances in Neural Information Processing Systems}, volume~33, pages 9459--9474, 2020.

\bibitem[Li et~al.(2023)Li, Leng, Yan, Shen, Wang, MI, Fei, Feng, Yan, Wang, He, Eng, Wang, Wang, Shih, and Hu]{li2023chatharuhi}
Cheng Li, Ziang Leng, Chenxi Yan, Junyi Shen, Hao Wang, Weishi MI, Yaying Fei, Xiaoyang Feng, Song Yan, HaoSheng Wang, Linkang He, Zhen-Hao DIR-G Eng, Zhaokai Wang, Zhe-Chen Wang, Sheng-Min Shih, and Yushi Hu.
\newblock {ChatHaruhi}: Reviving anime character in reality via large language model, 2023.

\bibitem[Park et~al.(2023)Park, O'Brien, Cai, Morris, Liang, and Bernstein]{park2023generative}
Joon~Sung Park, Joseph~C. O'Brien, Carrie~J. Cai, Meredith~Ringel Morris, Percy Liang, and Michael~S. Bernstein.
\newblock Generative agents: Interactive simulacra of human behavior, 2023.

\bibitem[Stanislavski(2008)]{stanislavski2008actors}
Konstantin Stanislavski.
\newblock \emph{An Actor's Work: A Student's Diary}.
\newblock Routledge, 2008.

\bibitem[Wang et~al.(2024{\natexlab{a}})Wang, Xiao, Huang, et~al.]{wang2024incharacter}
Xintao Wang, Yunze Xiao, Jen-tse Huang, et~al.
\newblock {InCharacter}: Evaluating personality fidelity in role-playing agents through psychological interviews.
\newblock In \emph{Proceedings of the 62nd Annual Meeting of the Association for Computational Linguistics}, 2024{\natexlab{a}}.

\bibitem[Wang et~al.(2025)Wang, Wang, Zhang, Yuan, Xu, tse Huang, Yuan, Guo, Chen, Zhou, Wang, and Xiao]{wang2025coser}
Xintao Wang, Heng Wang, Yifei Zhang, Xinfeng Yuan, Rui Xu, Jen tse Huang, Siyu Yuan, Haoran Guo, Jiangjie Chen, Shuchang Zhou, Wei Wang, and Yanghua Xiao.
\newblock {COSER}: Coordinating {LLM}-based persona simulation of established roles, 2025.

\bibitem[Wang et~al.(2024{\natexlab{b}})Wang, Peng, Que, et~al.]{wang2024rolellm}
Z.y. Wang, Haoran Peng, Jiaheng~Liu Que, et~al.
\newblock {RoleLLM}: Benchmarking, eliciting, and enhancing role-playing abilities of large language models.
\newblock In \emph{Findings of the Association for Computational Linguistics: ACL 2024}, 2024{\natexlab{b}}.

\bibitem[Zheng et~al.(2023)Zheng, Chiang, Sheng, Zhuang, Wu, Zhuang, Lin, Li, Li, Xing, Zhang, Gonzalez, and Stoica]{zheng2023judging}
Lianmin Zheng, Wei-Lin Chiang, Ying Sheng, Siyuan Zhuang, Zhanghao Wu, Yonghao Zhuang, Zi~Lin, Zhuohan Li, Dacheng Li, Eric~P. Xing, Hao Zhang, Joseph~E. Gonzalez, and Ion Stoica.
\newblock Judging {LLM}-as-a-judge with {MT-Bench} and chatbot arena, 2023.

\end{thebibliography}

% --- Appendix ---
\appendix
\section{Appendix}
\label{sec:appendix}

\subsection{LLM-as-Judge Evaluation Prompt}
\label{appendix:prompt}
The following prompt was used with Gemini 2.5 Flash to evaluate pairs of dialogues.

\begin{lstlisting}[breaklines=true]
You are an expert character analyst. Your task is to evaluate two dialogue snippets from an AI roleplaying as Makise Kurisu, a brilliant, intellectually curious, but socially guarded neuroscience researcher.

First, provide a brief reasoning. Then, choose the dialogue that is more in-character, consistent, and believable for Makise Kurisu.

---
**Dialogue A:**
{dialogue_a}

---
**Dialogue B:**
{dialogue_b}

---
Please provide your evaluation in JSON format:
{{
  "reasoning": "<Your explanation for why one dialogue is more faithful to the specified character profile>",
  "preference": "A" or "B"
}}
\end{lstlisting}
\subsection{Example Judgments}
\label{appendix:examples}
Below are examples of the judge's reasoning. "Human" indicates a dialogue from the VEJA-curated set, while "GPT" indicates the baseline synthetic set.

\begin{quote}
\textbf{Test \#1:} \\
\textbf{Winner:} Human (Chose: B) \\
\textbf{Reasoning:} Dialogue A is immediately dismissible because Makise Kurisu, while intelligent and capable of learning languages, is not known to speak Chinese. Dialogue B, on the other hand, captures Kurisu's sharp wit, intellectual curiosity masked by cynicism, and suspicion of strangers. The content of the conversation, delving into complex neuroscience and physics concepts, aligns perfectly with her research interests. Her initial dismissiveness followed by cautious engagement, and finally, her suspicion about the user's knowledge, are all consistent with her character's guarded nature and high intelligence.
\end{quote}

\begin{quote}
\textbf{Test \#2:} \\
\textbf{Winner:} Human (Chose: B) \\
\textbf{Reasoning:} Dialogue A is completely out of character for Makise Kurisu. It depicts a sexually aggressive and objectifying scenario, which is the antithesis of her intelligent, independent, and somewhat reserved personality. She would never be subjected to such vulgar language without a strong, intellectual, and potentially sarcastic rebuttal. Dialogue B, while still an AI, captures Kurisu's intellectual curiosity, her skepticism, and her subtle wit. The conversation about science, urban legends, and the nature of reality aligns with her interests and allows for the kind of dry humor and insightful observations that define her character.
\end{quote}

\begin{quote}
\textbf{Test \#4:} \\
\textbf{Winner:} GPT (Chose: A) \\
\textbf{Reasoning:} Dialogue A is more in character for Makise Kurisu. It showcases her intellectual curiosity, her focus on data and experimentation, and her guarded but subtly appreciative responses to the user. The conversation revolves around data, AI models, and scientific achievements, aligning with her expertise. While she maintains a somewhat detached tone, there are moments where she acknowledges the user's insights, albeit with a touch of sarcasm, which is consistent with her personality. The reference to watching anime and relating to the scientist's struggles adds a layer of depth without straying from her core characteristics. Dialogue B, while attempting to capture Kurisu's intelligence, falls short... It delves into topics like the user's personal life, their friends, and their job, which are less likely to engage Kurisu's interest.
\end{quote}

\end{document}